\documentclass[conference]{IEEEtran}
\IEEEoverridecommandlockouts
\usepackage{cite}
\usepackage{amsmath,amssymb,amsfonts}
\usepackage{algorithmic}
\usepackage{graphicx}
\usepackage{textcomp}
\usepackage{xcolor}
\usepackage{booktabs} 

\def\BibTeX{{\rm B\kern-.05em{\sc i\kern-.025em b}\kern-.08em
    T\kern-.1667em\lower.7ex\hbox{E}\kern-.125emX}}
\begin{document}


\title{GDAIP: A Graph-Based Domain Adaptive Framework for Individual Brain Parcellation}

\author{
\IEEEauthorblockN{1\textsuperscript{st} Jianfei Zhu}
\IEEEauthorblockA{\textit{Faculty of Computing} \\
\textit{Harbin Institute of Technology}\\
Harbin, China}
\and
\IEEEauthorblockN{2\textsuperscript{nd} Haiqi Zhu}
\IEEEauthorblockA{\textit{School of Medicine and Health} \\
\textit{Harbin Institute of Technology}\\
Harbin, China}
\and
\IEEEauthorblockN{3\textsuperscript{rd} Shaohui Liu}
\IEEEauthorblockA{\textit{Faculty of Computing} \\
\textit{Harbin Institute of Technology}\\
Harbin, China}
\and
\IEEEauthorblockN{4\textsuperscript{th} Feng Jiang}
\IEEEauthorblockA{\textit{School of Artificial Intelligence} \\
\textit{Nanjing University of Information Science and Technology}\\
Nanjing, China}
\and
\IEEEauthorblockN{5\textsuperscript{th} Baichun Wei}
\IEEEauthorblockA{\textit{School of Medicine and Health} \\
\textit{Harbin Institute of Technology}\\
Harbin, China \\
bcwei@hit.edu.cn}
\and
\IEEEauthorblockN{6\textsuperscript{th} Chuzhi Yi}
\IEEEauthorblockA{\textit{School of Medicine and Health} \\
\textit{Harbin Institute of Technology}\\
Harbin, China \\
chuzhiyi@hit.edu.cn}
}

\maketitle

\begin{abstract}
Recent deep learning approaches have shown promise in learning such individual brain parcellations from functional magnetic resonance imaging (fMRI). However, most existing methods assume consistent data distributions across domains and struggle with domain shifts inherent to real-world cross-dataset scenarios. To address this challenge, we proposed Graph Domain Adaptation for Individual Parcellation (GDAIP), a novel framework that integrates Graph Attention Networks (GAT) with Minimax Entropy (MME)-based domain adaptation. We construct cross-dataset brain graphs at both the group and individual levels. By leveraging semi-supervised training and adversarial optimization of the prediction entropy on unlabeled vertices from target brain graph, the reference atlas is adapted from the group-level brain graph to the individual brain graph, enabling individual parcellation under cross-dataset settings. We evaluated our method using parcellation visualization, Dice coefficient, and functional homogeneity. Experimental results demonstrate that GDAIP produces individual parcellations with topologically plausible boundaries, strong cross-session consistency, and ability of reflecting functional organization.
\end{abstract}


\section{Introduction}
Functional magnetic resonance imaging (fMRI) captures brain dynamics by measuring blood-oxygen-level-dependent (BOLD) signals \cite{canario2021review}. It has been widely applied in cognitive neuroscience \cite{fox2007spontaneous}, psychiatric diagnosis \cite{miranda2021systematic}, and neurological disease research \cite{teng2023brain}. Among various fMRI analysis strategies, node-based analysis plays a critical role owing to its efficiency in feature extraction and network modeling \cite{smith2011network}. This approach typically relies on a predefined atlas to divide the brain into functionally distinct regions (i.e., parcels), and then averages voxel- or vertex-level signals within each region to derive representative time series \cite{schaefer2018local}. These regional time series can then be used to construct functional connectivity (FC) networks or directly serve as features for downstream tasks such as classification, clustering, or prediction.

Node-based methods reduce the dimensionality of fMRI data and enhance signal stability, enabling the use of complex models such as Markov models \cite{vidaurre2018discovering} and deep neural networks \cite{sharma2023deep, ma2023multi}. The effectiveness of these methods heavily depends on the quality of the brain atlas, which plays a crucial role in capturing functional network topology and supporting accurate, biologically meaningful modeling \cite{moghimi2022evaluation}.

Various group-level brain atlases have been proposed based on fMRI data \cite{gordon2016generation, schaefer2018local, yeo2011organization, glasser2016multi}, which typically optimize for functional consistency across a population and produce stable, reusable parcellations. However, growing evidence highlights significant inter-individual variability in brain functional organization, even after precise anatomical alignment \cite{gordon2017individual, mueller2013individual}. This has led to increasing interest in individual brain parcellations, which aim to uncover subject-specific functional patterns while preserving group-level priors, thereby supporting fine-grained cognitive modeling and precision medicine \cite{li2025parcellation}.

In recent years, deep learning has emerged as a powerful tool for individual brain parcellation, owing to its ability to capture complex nonlinear patterns \cite{li2023application}. Several studies \cite{li2023computing, li2024ts, tian2022integrated, qiu2022unrevealing, ma2022bai} have proposed deep learning-based individual parcellation models trained on large-scale fMRI datasets and applied them to unseen individuals to infer subject-specific parcellations. However, most of these methods assume the same data distribution between training and testing, and overlook practical cross-dataset scenarios. In real-world settings, researchers often have access to only a limited number of subjects and must rely on models pre-trained on public datasets. Due to substantial differences across datasets—such as scanning protocols, acquisition parameters, and preprocessing pipelines—there exists a significant domain shift in fMRI data \cite{fang2025source, luppi2024systematic }, which can severely degrade model performance during transfer and even cause complete failure. Therefore, improving the cross-dataset generalization ability of individual parcellation models remains a key challenge for real-world application, and transfer learning is essential for addressing this issue.

In this study, we propose GDAIP, a novel individual parcellation framework based on GAT\cite{velivckovic2017graph} and MME\cite{saito2019semi} domain adaptation. Specifically, we construct graph representations of the brain using surface mesh topology and functional connectivity fingerprints as vertex features, generating both group-level and individual brain graphs across datasets. A semi-supervised GAT model is trained with a small set of labeled target-domain samples, and MME-based entropy regularization is introduced to enable feature alignment between group-level and individual representations, thereby improving cross-dataset transferability. We evaluate GDAP in terms of parcellation visualization intra- and inter-subject consistency, and functional homogeneity within regions. Our main contributions are as follows:

\begin{enumerate}
    \item To the best of our knowledge, we are the first to introduce transfer learning into individual brain parcellation to address the cross-dataset adaptation problem. The proposed GDAIP framework leverages group-level priors while simultaneously capturing individual functional organization, yielding reasonable individual brain parcellation.
    \item We designed a GAT-based individual brain parcellation model and integrate it with the MME domain adaptation mechanism.
    \item We conducted extensive experiments to evaluate the proposed method in terms of cross-session consistency, and functional homogeneity.
\end{enumerate}

\section{Method}
\subsection{Method Overview}
\begin{figure*}[ht]
\centerline{\includegraphics[width=0.85\textwidth,trim=0 80 0 100]{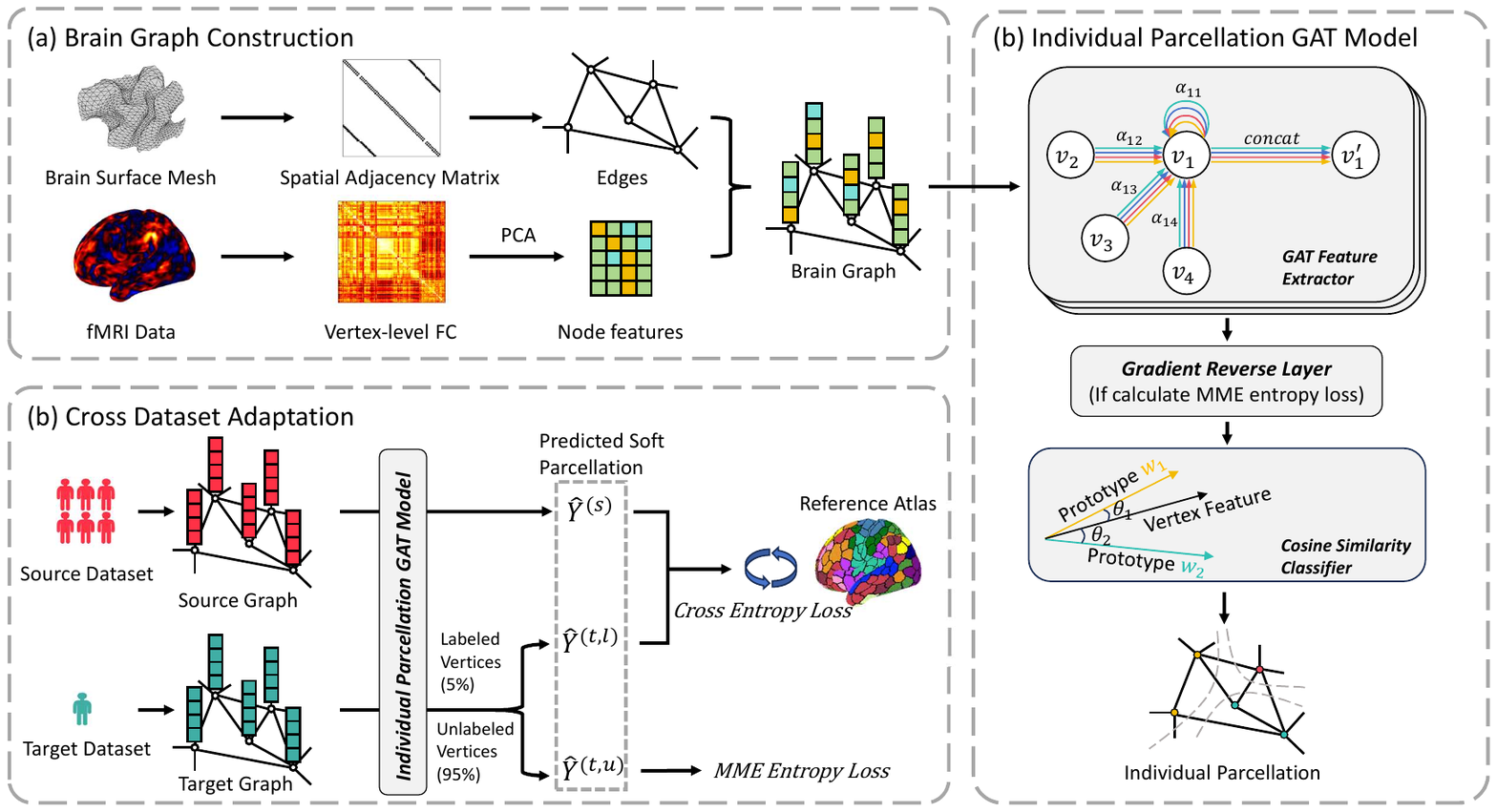}}
\caption{Overview of the GDAIP framework. (a) Brain graphs are constructed using surface-based adjacency and functional fingerprints from fMRI data. (b) The parcellation model includes a GAT-based feature extractor and a cosine similarity classifier, with a Gradient Reversal Layer enabling adversarial MME loss. (c) During training, graphs from source and target datasets are input to the individual parcellation model. Cross-entropy loss is computed on source and 5\% labeled target vertices, while MME loss is applied to the remaining unlabeled target vertices for cross-dataset adaptation.}
\label{fig:workflow}
\end{figure*}

This study aims to achieve cross-dataset individual adaptation of brain parcellation. We propose the GDAIP framework, which consists of three key steps: brain graph construction, graph self-attention modeling, and semi-supervised transfer training. The overall workflow is illustrated in Figure \ref{fig:workflow}.

We first construct brain graphs from fMRI data, where nodes correspond to cortical vertices and edges represent the adjacency relationships defined by the cortical surface mesh. Each node is associated with a feature vector derived from the Principal Component Analysis (PCA)-reduced functional fingerprint. A group-level brain graph is computed on the source dataset, while an individual brain graph is constructed for a single subject from the target dataset. We then apply a GAT model to extract contextual representations of the vertices by adaptively aggregating information from neighboring nodes via an attention mechanism. The final node representations are passed to a classifier based on cosine similarity to predict the corresponding brain parcel labels. 

To address the challenge of cross-dataset adaptation, we incorporate the MME transfer learning strategy, treating the group-level brain graph as the source domain and the individual brain graph as the target domain. The source brain graph use reference atlas as ground-truth. For the target brain graph, we compute the distance of each vertex to the atlas-defined boundaries, and label the top 5\% of vertices within each region of interest (ROI) (i.e., the core regions) using the same reference atlas. The remaining 95\% of vertices are treated as unlabeled. This setup establishes a semi-supervised training framework.

During training, two loss functions are jointly optimized. The supervised cross-entropy loss guides the model to learn group-level prior, while the adversarial MME entropy loss enhances the discriminability on unlabeled samples. With the combination of two losses, the model effectively leverages both group-level priors and individual-specific functional patterns, enabling individual brain parcellation across datasets.

\subsection{Construction of Brain Graph}
We adopt the GAT model to predict individual brain parcellations, which requires representing fMRI data in the form of a graph structure, referred to as a brain graph in this study. A graph is defined as $G={V,E}$, where $V$ denotes the set of vertices and E represents the set of edges. The edges can also be represented by an adjacency matrix $\mathbf{A}$. Separate brain graphs are constructed for the left and right hemispheres, and individual parcellations are generated independently for each hemisphere. The final parcellation is obtained by combining the results from both hemispheres.

We construct the adjacency matrix based on the fs\_LR\_32k cortical surface mesh. The mesh consists of triangular faces, each connecting three vertices. Vertices within the same triangle are considered connected, while vertices not sharing a triangle are considered unconnected. This mesh-based adjacency matrix captures the spatial neighborhood relationships on the cortical surface. Since all used fMRI data are aligned to the fs\_LR\_32k surface space, the adjacency matrix $\mathbf{A}\in R^{N\times N}$ is shared across subjects, where $N$ is the number of cortical vertices per hemisphere.

For each vertex, we compute a functional fingerprint and apply PCA for dimensionality reduction to obtain its feature vector. Specifically, we first compute a vertex-level FC matrix $\mathbf{C}\in R^{N\times N}$, where the $i$-th row $\mathbf{C}_i$ represents the FC between vertex $i$ and all other vertices, serving as the functional fingerprint of vertex. FC is calculated using Pearson correlation between vertex-wise time series: , where  and  are the time series of vertices $i$ and $j$, respectively. As the functional fingerprint has a high dimension, we apply PCA to reduce its dimensionality to and use the resulting vector as the input feature of each vertex.

To enable individual-level transfer, we compute the functional fingerprints for both the group-level and individual brain graphs. At the group level, vertex-level FC matrices are computed for all subjects in the source dataset, and the group-level functional fingerprint is obtained by averaging these matrices across all subjects. During PCA dimensionality reduction, we concatenate the group-level and individual FC matrices to ensure that their low-dimensional feature representations lie in the same feature space. Specifically:

\begin{equation}
    [\mathbf{F}^{(s)} \| \mathbf{F}^{(t)}] = PCA([\mathbf{C}^{(s)} \| \mathbf{C}^{(t)}])
\end{equation}

where $\mathbf{C}^{(s)}$ and $\mathbf{C}^{(t)}$ denote the group-level and individual FC matrices, respectively, and $\mathbf{F}^{(s)}, \mathbf{F}^{(t)} \in \mathbb{R}^{N \times d}$ are the resulting feature matrices. The symbol $\|$ represents matrix concatenation.

\subsection{GAT model for Individual Parcellation}
In this work, our individual parcellation model consists of a GAT feature extractor and a cosine similarity-based classifier. GAT effectively captures the local structural features among vertices by introducing an attention mechanism that assigns different importance weights to neighboring nodes, enabling adaptive information aggregation. Furthermore, GAT employs a multi-head attention mechanism that models vertex relationships from multiple subspaces, enhancing both the expressiveness and stability of the model. 
In our implementation, we use a three-layer GAT as the feature extractor. Each layer contains 4 attention heads with an output dimension of 50 per head, resulting in a total feature dimension of 200 for each layer.
Following the original MME framework, we employ a cosine similarity-based classifier to enhance feature discriminability and generalization. The classifier performs L2 normalization on input features and computes cosine similarity between features and class prototypes using a bias-free linear layer. A temperature scaling factor is applied to control the sharpness of the SoftMax output. The predicted probability for vertex i belonging to class r is given by:

\begin{equation}
    \widehat{\mathbf{Y}}_{i,r} = \frac{\exp\left(\mathbf{w}_r^{\top} \widetilde{\mathbf{F}}_i / \tau \right)}
    {\sum_{j=1}^{N_{\text{roi}}} \exp\left( \mathbf{w}_j^{\top} \widetilde{\mathbf{F}}_j / \tau \right)}
\end{equation}

where $\widetilde{\mathbf{F}}_i \in \mathbb{R}^d$ denotes the feature extractor’s output of the $i$-th vertex, $\mathbf{w}_r \in \mathbb{R}^d$ is the prototype vector (i.e., weight vector) for class $r$, $\tau = 0.05$ is the temperature parameter, and $N_{\text{roi}}$ denotes the total number of classes (i.e., ROIs). All prototype vectors $\{\mathbf{w}_r\}_{r=1}^{N_{\text{roi}}}$ are learnable parameters optimized via backpropagation during training.

\subsection{Self-supervised Training and Domain adaptation}
We adopt a semi-supervised transfer learning strategy to train the individual parcellation GAT model. The Schaefer400 atlas \cite{schaefer2018local} is used to provide labels for the vertices from source domain (group-level graph). In the target domain (individual graph), a small number of vertices near the center of each ROI (as defined by the reference atlas) are selected as semi-supervised samples. The GAT model is trained using all available labeled vertices from both domains. To address the distribution shift between group-level and individual features, we employ the MME semi-supervised transfer learning method. This allows the model to effectively leverage group-level priors while adapting to individual-specific characteristics using only a small number of labeled nodes, thereby enabling accurate individual brain parcellation.

\subsubsection{Atlas-guided Semi-supervised Training}
Since the source domain features are derived from group-averaged FC matrix, and the reference atlas reflects group-level clustering priors, it is reasonable to use the reference atlas as vertex labels for the group-level brain graph. Semi-supervised transfer learning requires a small number of labeled samples in the target domain. Thus, we assume that differences between individual and reference parcellations primarily occur near the ROI boundaries, while the central regions of each ROI remain consistent and reliable. Therefore, we extract the most representative core region within each ROI from the reference atlas, and use the corresponding vertices in the individual brain graph as labeled samples (also labeled using reference atlas) for semi-supervised learning.

Specifically, based on the vertex adjacency of the brain graph, the core region of each ROI is identified as follows: (1) Using the adjacency matrix, we identify boundary vertices—those within an ROI that are adjacent to vertices belonging to other ROIs. (2) For all vertices in the ROI, we compute their shortest-path distance to the set of boundary vertices using a graph-based distance algorithm. (3) The vertices are then ranked by distance, and the top 5\% farthest from the boundary are selected as the ROI’s core region.

Cross-entropy loss is optimized over all labeled samples, including all vertices in the source domain and the core-region vertices in the target domain:

\begin{equation}
\begin{aligned}
L_{\text{cls}} = & - \frac{1}{|V^{(s)}|} \sum_{i \in V^{(s)}} \sum_{r=1}^{N_{\text{roi}}} \mathbb{1}[y^{(s)}_i = r] \log \left( \widehat{\mathbf{Y}}^{(s)}_{i,r} \right) \\
& - \frac{1}{|V^{(t,l)}|} \sum_{i \in V^{(t,l)}} \sum_{r=1}^{N_{\text{roi}}} \mathbb{1}[y^{(t,l)}_i = r] \log \left( \widehat{\mathbf{Y}}^{(t,l)}_{i,r} \right)
\end{aligned}
\end{equation}

Where $V^{(s)}$ denote the set of vertices in the source domain, and $V^{(t,l)}$ denote the set of labeled vertices in the target domain. The corresponding label vectors are $\mathbf{y}^{(s)} \in \mathbb{R}^{|V^{(s)}| \times 1}$ and $\mathbf{y}^{(t,l)} \in \mathbb{R}^{|V^{(t,l)}| \times 1}$, where each element represents the parcellation label of a vertex. The predicted probability that the $i$-th vertex in the source domain belongs to the $r$-th brain region is denoted as $\widehat{\mathbf{Y}}^{(s)}_{i,r}$, and similarly, $\widehat{\mathbf{Y}}^{(t,l)}_{i,r}$ denotes the predicted probability for the labeled vertices in the target domain. The indicator function $\mathbb{1}_{[\cdot]}$ returns 1 if the condition is true, and 0 otherwise.

\subsubsection{Domain Adaptation based on MME}
Since the number of labeled vertices in the source domain is significantly larger than in the target domain, a classifier trained solely with the above cross-entropy loss tends to be biased toward the source domain, leading to poor generalization on the unlabeled vertices in the target domain. To address this issue, we introduce the MME strategy as our transfer learning method.

MME is a classical domain adaptation approach that enhances feature discriminability by introducing adversarial optimization on the prediction entropy of unlabeled target-domain samples. This strategy can be naturally adapted to graph neural network training \cite{xiao2024semi}.

For unlabeled vertices in the target domain, we introduce the following entropy regularization loss:

\begin{equation}
L_{\text{ent}} = -\frac{1}{|V^{(t,u)}|} \sum_{i \in V^{(t,u)}} \sum_{r=1}^{N_{\text{roi}}} 
\widehat{\mathbf{Y}}^{(t,u)}_{i,r} \log \left( \widehat{\mathbf{Y}}^{(t,u)}_{i,r} \right)
\end{equation}

Here, $V^{(t,u)}$ denotes the set of unlabeled vertices in the target domain, and 
$\widehat{\mathbf{Y}}^{(t,u)}_{i,r}$ represents the predicted probability that the $i$-th 
unlabeled vertex belongs to the $r$-th brain region. The entropy term reflects the model's 
uncertainty in classifying unlabeled vertices: higher entropy indicates more ambiguous predictions 
across classes, whereas lower entropy implies stronger discriminative capability.

The MME optimization process can be interpreted as an adversarial training mechanism consisting of two stages. In the first stage, the feature extractor is fixed, and the classifier is optimized to maximize the entropy of predictions on the unlabeled target-domain vertices. This encourages the classifier to produce more “confused” outputs, thereby forcing class prototypes to move closer to the center of target-domain features. In the second stage, the classifier is fixed, and the feature extractor is updated to minimize the entropy, which drives the target-domain features to cluster around specific prototypes and enhances class separability.

In practice, it is not necessary to alternate these two steps explicitly. Instead, we adopt a Gradient Reversal Layer (GRL) to enable joint adversarial optimization. The GRL behaves as an identity mapping during the forward pass but reverses the gradient sign during backpropagation. This design allows the classifier to learn to maximize entropy, while the feature extractor is implicitly trained to minimize entropy. This mechanism avoids the computational overhead of alternating optimization and promotes alignment between the source and target feature spaces, ultimately enabling efficient and robust individual adaptation of brain parcellation across datasets.

\subsection{Implementation Details}
During model training, we adopt a full-graph training strategy, where all vertices from both the source and target brain graphs are used in each training step. This setting allows for maximum utilization of the global graph structure. The training process lasts for 4000 steps. The initial learning rate is set to 0.01 and is reduced by half after every 1000 steps.

We use stochastic gradient descent (SGD) with momentum as the optimizer, with the momentum coefficient set to 0.9 and the weight decay parameter set to 0.0005. The model is trained on a NVIDIA GeForce 4090 with 24G video memory.

\subsection{Validation Metrics}
To evaluate the consistency of individual brain parcellation, we use the Dice coefficient to measure the overlap of ROIs between parcellations. Intra-subject consistency reflects the stability of parcellations across sessions for the same subject, while inter-subject consistency reflects variability across different individuals. For each ROI, we compute the Dice coefficient and average the results. A significantly higher intra-subject Dice score compared to inter-subject indicates that the model can reliably preserve individual-specific functional organization across sessions, rather than producing generic or population-level patterns. We use an independent two-sample t-test to assess the statistical significance of this difference, which serves as an indicator of the model’s ability to capture individual variability.

To assess how well the parcellation reflects functional coherence, we calculate functional homogeneity by measuring the similarity of BOLD time series within each region. Specifically, we compute the average Pearson correlation among all vertex time series within each ROI, then average across all regions. Higher homogeneity suggests that the parcellation delineates functionally coherent regions, supporting the validity of the individualized results.

\section{Experiment Results and Discussions}
\subsection{Dataset and Preprocessing}
In this study, we evaluated the proposed method on two public datasets: the HCP Young Adult dataset \cite{van2013wu} and the NKI Rockland Sample dataset \cite{nooner2012nki}. These datasets were used in a cross-domain setting, where each dataset served as a source domain to adapt to the other as the target domain, allowing us to assess the cross-dataset adaptability of our approach. Only resting-state fMRI (rs-fMRI) data were used for model training and validation.

The HCP Young Adult dataset contains rs-fMRI data from approximately 1,200 subjects. Most participants underwent four rs-fMRI sessions acquired over two days: REST1\_LR, REST1\_RL, REST2\_LR, and REST2\_RL. Each session consisted of 1,200 time points (TR = 0.72 s). 

The NKI Rockland Sample dataset includes rs-fMRI data from approximately 1,300 participants. Each subject was scanned in two rs-fMRI sessions with different acquisition protocols: one with high temporal resolution (TR = 0.645 s, voxel size = 3 mm) and the other with high spatial resolution (TR = 1.4 s, voxel size = 2 mm).

In this study, we randomly selected 400 subjects from each dataset to validate our method. For the HCP dataset, we used the minimally preprocessed data provided by HCP \cite{glasser2013minimal} (i.e., rfMRI\_REST1\_LR\_Atlas.dtseries.nii), where cortical data were already mapped to the standard fs\_LR\_32k surface space. For the NKI dataset, we utilized fMRIprep \cite{Esteban2019fMRIPrep} with default parameters to preprocess the MRI data. After preprocessing, the fMRI data are projected onto the fs\_LR\_32k surface space using the volume-to-surface mapping tool provided by HCP Workbench \cite{marcus2011informatics}.

\subsection{Competing Methods}
Since GDAIP was guided by a reference parcellation during training, the resulting individual parcellations maintained a one-to-one correspondence with the reference, enabling region-level alignment. Therefore, we selected baseline methods that also support this one-to-one correspondence for fair comparison. Three individual brain parcellation methods were included in our comparison.

First, we considered two learning-based approaches: an MLP-based binary classification method (referred to as MLPBC in this paper) \cite{tian2022integrated}, and a GCN-based node classification method (MSGCN) \cite{qiu2022unrevealing}. Both methods use the reference atlas to provide labels, and train models on group-level data, which are then applied to individual data to generate subject-specific parcellations. To ensure fairness in comparison, we fine-tuned both models (denoted as MLPBC-FT and MSGCN-FT, respectively) using the same setting as our method: 5\% of the vertices of the individual brain graph were used as labeled samples for fine-tuning.

In addition, we included a reference-guided optimization-based method (AGP)\cite{li2022atlas}. AGP initializes seed regions based on the reference atlas and iteratively assigns unlabeled vertices to the neighboring parcel with the highest functional similarity. The resulting individual parcellation also maintains one-to-one correspondence with the reference and is generated independently for each subject without requiring cross-dataset transfer. Comparing with AGP highlighted the advantages of learning-based methods over optimization-based methods, and further demonstrated the necessity of cross-dataset transfer adaptation.

\subsection{Visualization of Individual Parcellations}

\begin{figure*}[htbp]
\centerline{\includegraphics[width=0.95\textwidth,trim=0 90 0 100]{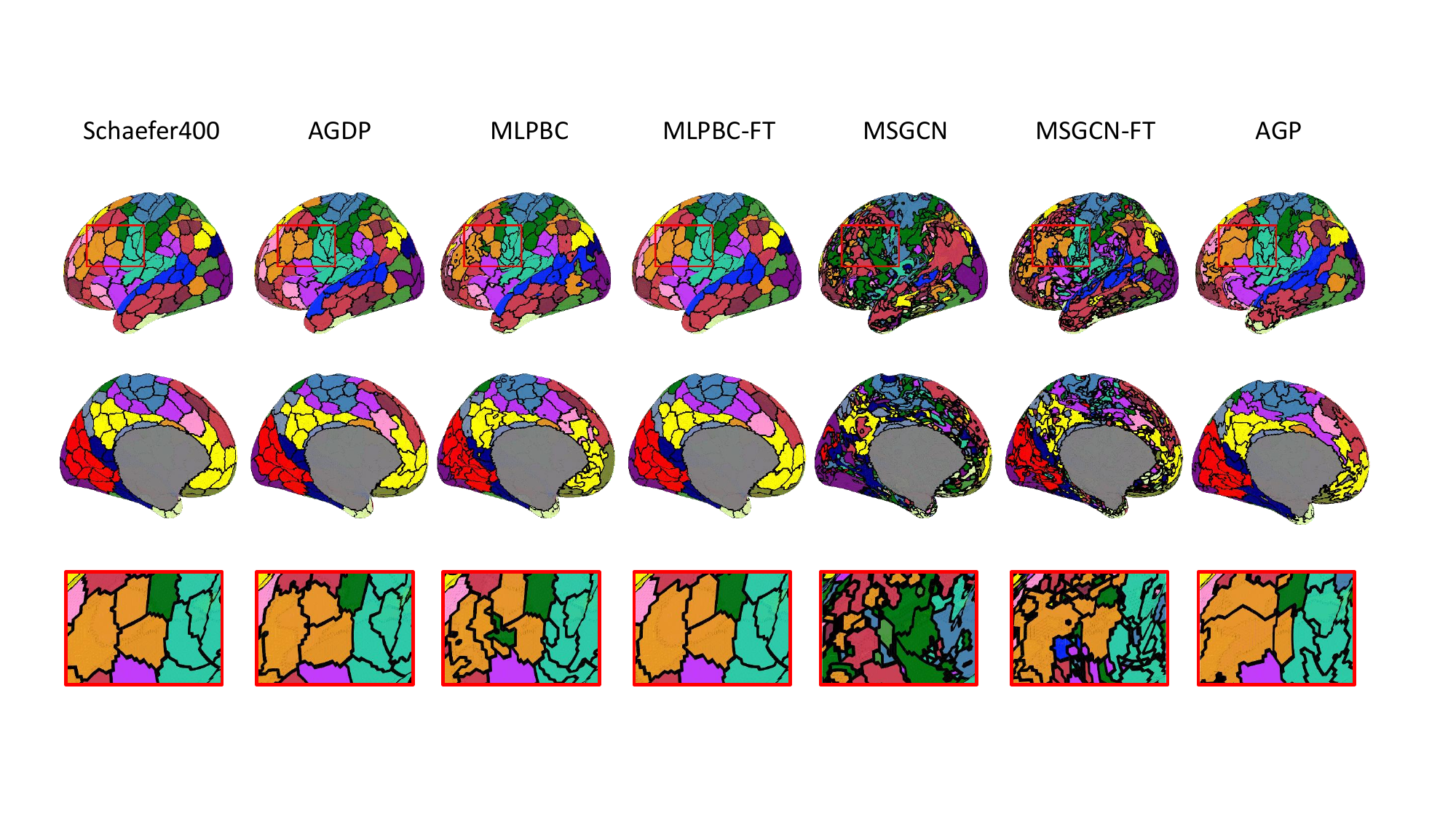}}
\caption{Visualization of the reference Schaefer400 atlas and individual parcellations from different methods. Parcellations are generated for subject 100408 from the HCP dataset, using the NKI dataset as the source domain.}
\label{fig:vis}
\end{figure*}

We visualized the Schaefer400 reference atlas and the individual parcellations generated by GDAIP and three baseline methods. Figure \ref{fig:vis} presents the individual parcellation results for subject 100408 from the HCP dataset, adapted from the NKI dataset. As shown in the results, our proposed GDAIP method generated parcellations that were notably different from the reference atlas, while still preserving the relative spatial organization and continuity of ROI shapes. The resulting parcels were mostly convex, topologically simple, and exhibited smooth boundaries, which aligned well with the characteristics of the Schaefer400 atlas. This demonstrated that GDAIP effectively captured individual-specific differences during cross-dataset adaptation, while simultaneously leveraging group-level priors to produce reasonable brain parcellations.

The individual parcellations generated by MLPBC preserved the relative spatial arrangement of ROIs defined in the reference atlas, but exhibited rugged boundaries and discontinuous topological structures. After fine-tuning, the parcellation results became overly similar to the reference atlas, failing to reflect subject-specific variations. This might be attributed to the way vertex features were constructed in MLPBC: each vertex's input was based on correlation with atlas-defined ROIs, which may lead the model to learn a shortcut by exploiting identity information (e.g., a vertex correlates perfectly with itself), resulting in trivial reproduction of the atlas.

The results of MSGCN indicated a failure in cross-dataset individual parcellation adaptation. Some parcels were topologically complex and discontinuous, often fragmenting into multiple small, isolated patches. In addition, the relative spatial organization of ROIs poorly aligned with that of the reference atlas. Although the fine-tuned MSGCN showed some improvement in the spatial positioning of brain regions, the overall topological structure remained suboptimal. This suggested that the model trained on the source domain lacked sufficient generalization ability across datasets. Even after fine-tuning, the model failed to produce reasonable parcellations, indicating that shallow adaptation using a limited number of labeled vertices was insufficient for correcting the distributional shift between source and target domains.

The parcellation results produced by AGP showed greater divergence from the reference atlas compared to those generated by GDAIP, and the resulting individual parcellation boundaries tended to be more jagged and irregular. As an optimization-based method, AGP does not suffer from cross-dataset transfer issues like learning-based methods. However, it may lack the ability to incorporate group-level priors, which leads to possible overfitting to individual fMRI data. This could explain the more fragmented and serrated ROI boundaries, as well as the greater deviation from the reference atlas.

\subsection{Results of Dice Coefficient}
We presented the Dice coefficients of individual parcellations generated by GDAIP and baseline methods in Figure 3. The figure displays both intra-subject and inter-subject Dice coefficients for each method, along with significance markers indicating statistical differences between the two. Subfigure (a) presents the results of the experimental setting where the NKI dataset serves as the source domain and the HCP dataset as the target domain. Subfigure (b) shows the reverse setting, where the HCP dataset is used as the source domain and the NKI dataset as the target domain.

\begin{figure}[ht]
\centerline{\includegraphics[width=0.35\textwidth,trim=0 0 0 0]{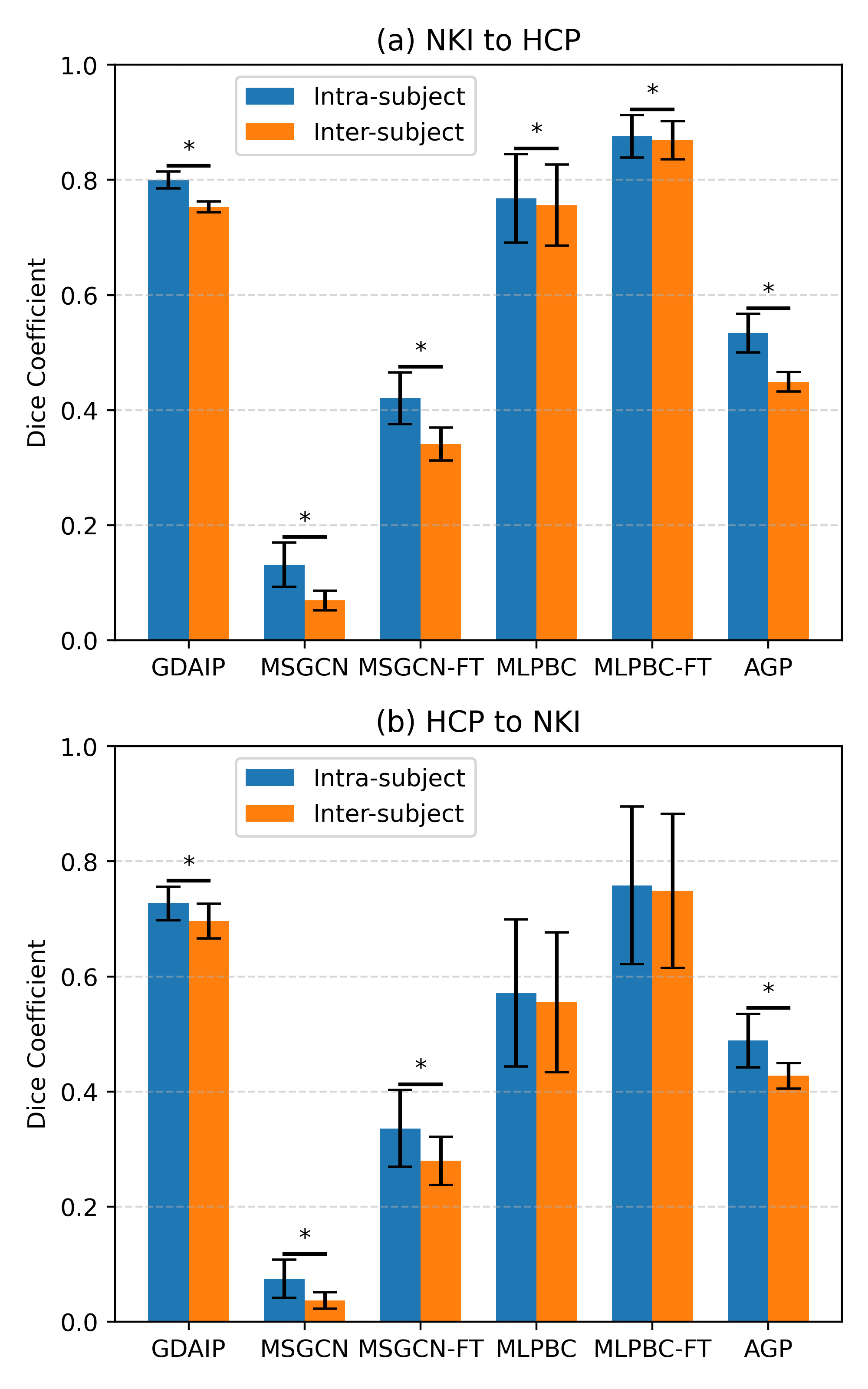}}
\caption{ Results of intra- and inter-subject Dice coefficient of individual parcellations across different methods.}
\label{fig:dice}
\end{figure}

From the results, we observed that our GDAIP method achieved high intra-subject Dice scores, second only to MLPBC-FT. This indicated that GDAIP can produce consistent individual parcellations across different fMRI sessions of the same subject. As shown in the previous visualization, the parcellations produced by MLPBC-FT closely resembled the reference atlas, which may explain its elevated intra-subject Dice scores. In contrast, MSGCN, MSGCN-FT, and AGP showed relatively low intra-subject Dice values, suggesting substantial variation in parcellations across sessions of the same subject. 

With the exception of MLPBC and MLPBC-FT under the HCP-to-NKI experiment setting, all individual parcellation methods yielded significantly higher intra-subject Dice coefficients compared to inter-subject Dice. This suggested that GDAIP, MSGCN, MSGCN-FT, and AGP were all capable of capturing individual differences to some extent. However, the individual differences captured by GDAIP were smaller than those of the other three methods. Combined with the intra-subject Dice results and visual comparisons, this implied that the other methods may have overfitted to session-specific noise, while GDAP better leveraged group-level priors to maintain stable and biologically meaningful parcellations.

In summary, the Dice coefficient offers two criteria for evaluating individual parcellation: (1) a high intra-subject Dice score to ensure reproducibility; and (2) a significantly lower inter-subject Dice score to reflect subject-specific variability. GDAIP outperformed baseline methods in balancing both aspects, demonstrating better consistency and generalizability in individual brain parcellation.

\subsection{Results of Functional Homogeneity}
The functional homogeneity results are shown in Figure 4. We compared the functional homogeneity of individual parcellations generated by each method against that of the reference atlas, with statistical significance indicated in the figure. Similar to the Dice coefficient results, Subfigures (a) and (b) present the results under the NKI-to-HCP and HCP-to-NKI experimental settings, respectively.

\begin{figure}[ht]
\centerline{\includegraphics[width=0.35\textwidth,trim=0 0 0 0]{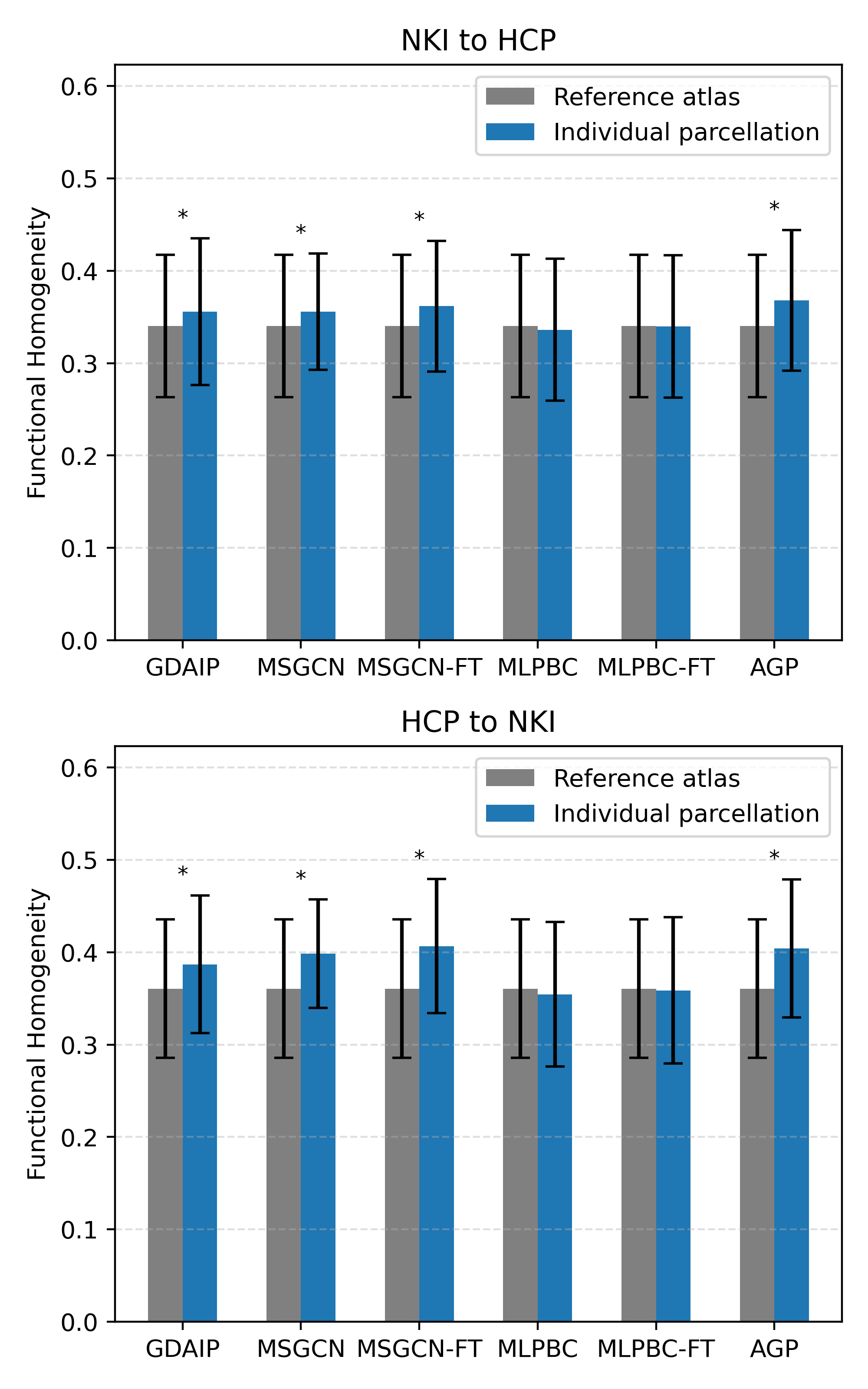}}
\caption{Comparison of functional homogeneity of individual parcellations and the reference atlas.}
\label{fig:func}
\end{figure}

As shown in the results, the individual parcellations produced by GDAIP exhibited significantly higher functional homogeneity than the reference atlas, indicating that GDAP-generated regions were more consistent with the underlying individual functional topology. In contrast, MLPBC and MLPBC-FT did not show significant improvements over the reference atlas in either adaptation setting, suggesting that they failed to capture individual-specific functional characteristics in cross-dataset scenarios.

MSGCN and MSGCN-FT achieved higher homogeneity scores than GDAIP; however, as observed in the visualization results, these methods failed to produce topologically plausible parcellations. Their inflated homogeneity scores may result from fragmented, overly fine-grained subregions, which artificially increased within-region time series similarity.

AGP also yielded strong functional homogeneity. As an optimization-based method, AGP independently computed individual parcellations on single-subject fMRI data without relying on group-level priors, which gave it an advantage in preserving within-region coherence.

\section{Conclusion}
This work introduced transfer learning to address the challenge of cross-dataset adaptation in deep learning-based individual brain parcellation. We proposed GDAIP, a novel framework that constructed both group-level and individual brain graphs and adapted the reference atlas from the group-level brain graph to the individual brain graph. We evaluate the parcellation performance from four perspectives: visualization, Dice coefficient, and functional homogeneity. Experimental results showed that existing learning-based methods fail to produce reliable individualized parcellations under cross-dataset conditions. In contrast, GDAIP successfully generated individualized parcellations with topologically coherent structures, as evidenced by qualitative visualizations. Evaluation using the Dice coefficient demonstrated strong across-session stability, while the significantly higher functional homogeneity compared to the reference atlas indicated better alignment with individual-specific functional organization. These results demonstrated that GDAIP effectively leveraged group-level priors while generating individualized parcellations that reflect subject-specific functional characteristics.

\bibliographystyle{IEEEtran}
\bibliography{refs}

\begin{thebibliography}{10}
\providecommand{\url}[1]{#1}
\csname url@samestyle\endcsname
\providecommand{\newblock}{\relax}
\providecommand{\bibinfo}[2]{#2}
\providecommand{\BIBentrySTDinterwordspacing}{\spaceskip=0pt\relax}
\providecommand{\BIBentryALTinterwordstretchfactor}{4}
\providecommand{\BIBentryALTinterwordspacing}{\spaceskip=\fontdimen2\font plus
\BIBentryALTinterwordstretchfactor\fontdimen3\font minus \fontdimen4\font\relax}
\providecommand{\BIBforeignlanguage}[2]{{%
\expandafter\ifx\csname l@#1\endcsname\relax
\typeout{** WARNING: IEEEtran.bst: No hyphenation pattern has been}%
\typeout{** loaded for the language `#1'. Using the pattern for}%
\typeout{** the default language instead.}%
\else
\language=\csname l@#1\endcsname
\fi
#2}}
\providecommand{\BIBdecl}{\relax}
\BIBdecl

\bibitem{canario2021review}
E.~Canario, D.~Chen, and B.~Biswal, ``A review of resting-state fmri and its use to examine psychiatric disorders,'' \emph{Psychoradiology}, vol.~1, no.~1, pp. 42--53, 2021.

\bibitem{fox2007spontaneous}
M.~D. Fox and M.~E. Raichle, ``Spontaneous fluctuations in brain activity observed with functional magnetic resonance imaging,'' \emph{Nature reviews neuroscience}, vol.~8, no.~9, pp. 700--711, 2007.

\bibitem{miranda2021systematic}
L.~Miranda, R.~Paul, B.~P{\"u}tz, N.~Koutsouleris, and B.~M{\"u}ller-Myhsok, ``Systematic review of functional mri applications for psychiatric disease subtyping,'' \emph{Frontiers in Psychiatry}, vol.~12, p. 665536, 2021.

\bibitem{teng2023brain}
J.~Teng, C.~Mi, J.~Shi, and N.~Li, ``Brain disease research based on functional magnetic resonance imaging data and machine learning: a review,'' \emph{Frontiers in Neuroscience}, vol.~17, p. 1227491, 2023.

\bibitem{smith2011network}
S.~M. Smith, K.~L. Miller, G.~Salimi-Khorshidi, M.~Webster, C.~F. Beckmann, T.~E. Nichols, J.~D. Ramsey, and M.~W. Woolrich, ``Network modelling methods for fmri,'' \emph{Neuroimage}, vol.~54, no.~2, pp. 875--891, 2011.

\bibitem{schaefer2018local}
A.~Schaefer, R.~Kong, E.~M. Gordon, T.~O. Laumann, X.-N. Zuo, A.~J. Holmes, S.~B. Eickhoff, and B.~T. Yeo, ``Local-global parcellation of the human cerebral cortex from intrinsic functional connectivity mri,'' \emph{Cerebral cortex}, vol.~28, no.~9, pp. 3095--3114, 2018.

\bibitem{vidaurre2018discovering}
D.~Vidaurre, R.~Abeysuriya, R.~Becker, A.~J. Quinn, F.~Alfaro-Almagro, S.~M. Smith, and M.~W. Woolrich, ``Discovering dynamic brain networks from big data in rest and task,'' \emph{NeuroImage}, vol. 180, pp. 646--656, 2018.

\bibitem{sharma2023deep}
R.~Sharma, T.~Goel, M.~Tanveer, C.-T. Lin, and R.~Murugan, ``Deep-learning-based diagnosis and prognosis of alzheimer’s disease: a comprehensive review,'' \emph{IEEE Transactions on Cognitive and Developmental Systems}, vol.~15, no.~3, pp. 1123--1138, 2023.

\bibitem{ma2023multi}
Y.~Ma, Q.~Wang, L.~Cao, L.~Li, C.~Zhang, L.~Qiao, and M.~Liu, ``Multi-scale dynamic graph learning for brain disorder detection with functional mri,'' \emph{IEEE Transactions on Neural Systems and Rehabilitation Engineering}, vol.~31, pp. 3501--3512, 2023.

\bibitem{moghimi2022evaluation}
P.~Moghimi, A.~T. Dang, Q.~Do, T.~I. Netoff, K.~O. Lim, and G.~Atluri, ``Evaluation of functional mri-based human brain parcellation: a review,'' \emph{Journal of neurophysiology}, vol. 128, no.~1, pp. 197--217, 2022.

\bibitem{gordon2016generation}
E.~M. Gordon, T.~O. Laumann, B.~Adeyemo, J.~F. Huckins, W.~M. Kelley, and S.~E. Petersen, ``Generation and evaluation of a cortical area parcellation from resting-state correlations,'' \emph{Cerebral cortex}, vol.~26, no.~1, pp. 288--303, 2016.

\bibitem{yeo2011organization}
B.~T. Yeo, F.~M. Krienen, J.~Sepulcre, M.~R. Sabuncu, D.~Lashkari, M.~Hollinshead, J.~L. Roffman, J.~W. Smoller, L.~Z{\"o}llei, J.~R. Polimeni \emph{et~al.}, ``The organization of the human cerebral cortex estimated by intrinsic functional connectivity,'' \emph{Journal of neurophysiology}, 2011.

\bibitem{glasser2016multi}
M.~F. Glasser, T.~S. Coalson, E.~C. Robinson, C.~D. Hacker, J.~Harwell, E.~Yacoub, K.~Ugurbil, J.~Andersson, C.~F. Beckmann, M.~Jenkinson \emph{et~al.}, ``A multi-modal parcellation of human cerebral cortex,'' \emph{Nature}, vol. 536, no. 7615, pp. 171--178, 2016.

\bibitem{gordon2017individual}
E.~M. Gordon, T.~O. Laumann, B.~Adeyemo, and S.~E. Petersen, ``Individual variability of the system-level organization of the human brain,'' \emph{Cerebral cortex}, vol.~27, no.~1, pp. 386--399, 2017.

\bibitem{mueller2013individual}
S.~Mueller, D.~Wang, M.~D. Fox, B.~T. Yeo, J.~Sepulcre, M.~R. Sabuncu, R.~Shafee, J.~Lu, and H.~Liu, ``Individual variability in functional connectivity architecture of the human brain,'' \emph{Neuron}, vol.~77, no.~3, pp. 586--595, 2013.

\bibitem{li2025parcellation}
C.~Li, S.~Yu, and Y.~Cui, ``Parcellation of individual brains: From group level atlas to precise mapping,'' \emph{Neuroscience \& Biobehavioral Reviews}, p. 106172, 2025.

\bibitem{li2023application}
Y.~Li, X.~Chen, Q.~Ling, Z.~He, and A.~Liu, ``Application of deep learning in fmri-based human brain parcellation: a review,'' \emph{Measurement Science and Technology}, vol.~35, no.~3, p. 032001, 2023.

\bibitem{li2023computing}
H.~Li, D.~Srinivasan, C.~Zhuo, Z.~Cui, R.~E. Gur, R.~C. Gur, D.~J. Oathes, C.~Davatzikos, T.~D. Satterthwaite, and Y.~Fan, ``Computing personalized brain functional networks from fmri using self-supervised deep learning,'' \emph{Medical Image Analysis}, vol.~85, p. 102756, 2023.

\bibitem{li2024ts}
C.~Li, Y.~Lu, S.~Yu, and Y.~Cui, ``Ts-ai: A deep learning pipeline for multimodal subject-specific parcellation with task contrasts synthesis,'' \emph{Medical Image Analysis}, vol.~97, p. 103297, 2024.

\bibitem{tian2022integrated}
X.~Tian, Y.~Chen, P.~Majka, D.~Szczupak, Y.~S. Perl, C.~C.-C. Yen, C.~Tong, F.~Feng, H.~Jiang, D.~Glen \emph{et~al.}, ``An integrated resource for functional and structural connectivity of the marmoset brain,'' \emph{Nature Communications}, vol.~13, no.~1, p. 7416, 2022.

\bibitem{qiu2022unrevealing}
W.~Qiu, L.~Ma, T.~Jiang, and Y.~Zhang, ``Unrevealing reliable cortical parcellation of individual brains using resting-state functional magnetic resonance imaging and masked graph convolutions,'' \emph{Frontiers in neuroscience}, vol.~16, p. 838347, 2022.

\bibitem{ma2022bai}
L.~Ma, Y.~Zhang, H.~Zhang, L.~Cheng, Z.~Yang, Y.~Lu, W.~Shi, W.~Li, J.~Zhuo, J.~Wang \emph{et~al.}, ``Bai-net: Individualized anatomical cerebral cartography using graph neural network,'' \emph{IEEE transactions on neural networks and learning systems}, vol.~35, no.~6, pp. 7446--7457, 2022.

\bibitem{fang2025source}
Y.~Fang, J.~Wu, Q.~Wang, S.~Qiu, A.~Bozoki, and M.~Liu, ``Source-free collaborative domain adaptation via multi-perspective feature enrichment for functional mri analysis,'' \emph{Pattern recognition}, vol. 157, p. 110912, 2025.

\bibitem{luppi2024systematic}
A.~I. Luppi, H.~M. Gellersen, Z.-Q. Liu, A.~R. Peattie, A.~E. Manktelow, R.~Adapa, A.~M. Owen, L.~Naci, D.~K. Menon, S.~I. Dimitriadis \emph{et~al.}, ``Systematic evaluation of fmri data-processing pipelines for consistent functional connectomics,'' \emph{Nature Communications}, vol.~15, no.~1, p. 4745, 2024.

\bibitem{velivckovic2017graph}
P.~Veli{\v{c}}kovi{\'c}, G.~Cucurull, A.~Casanova, A.~Romero, P.~Lio, and Y.~Bengio, ``Graph attention networks,'' \emph{arXiv preprint arXiv:1710.10903}, 2017.

\bibitem{saito2019semi}
K.~Saito, D.~Kim, S.~Sclaroff, T.~Darrell, and K.~Saenko, ``Semi-supervised domain adaptation via minimax entropy,'' in \emph{Proceedings of the IEEE/CVF international conference on computer vision}, 2019, pp. 8050--8058.

\bibitem{xiao2024semi}
J.~Xiao, Q.~Dai, X.~Shen, X.~Xie, J.~Dai, J.~Lam, and K.-W. Kwok, ``Semi-supervised domain adaptation on graphs with contrastive learning and minimax entropy,'' \emph{Neurocomputing}, vol. 580, p. 127469, 2024.

\bibitem{van2013wu}
D.~C. Van~Essen, S.~M. Smith, D.~M. Barch, T.~E. Behrens, E.~Yacoub, K.~Ugurbil, W.-M.~H. Consortium \emph{et~al.}, ``The wu-minn human connectome project: an overview,'' \emph{Neuroimage}, vol.~80, pp. 62--79, 2013.

\bibitem{nooner2012nki}
K.~B. Nooner, S.~J. Colcombe, R.~H. Tobe, M.~Mennes, M.~M. Benedict, A.~L. Moreno, L.~J. Panek, S.~Brown, S.~T. Zavitz, Q.~Li \emph{et~al.}, ``The nki-rockland sample: a model for accelerating the pace of discovery science in psychiatry,'' \emph{Frontiers in neuroscience}, vol.~6, p. 152, 2012.

\bibitem{glasser2013minimal}
M.~F. Glasser, S.~N. Sotiropoulos, J.~A. Wilson, T.~S. Coalson, B.~Fischl, J.~L. Andersson, J.~Xu, S.~Jbabdi, M.~Webster, J.~R. Polimeni \emph{et~al.}, ``The minimal preprocessing pipelines for the human connectome project,'' \emph{Neuroimage}, vol.~80, pp. 105--124, 2013.

\bibitem{Esteban2019fMRIPrep}
O.~Esteban, C.~J. Markiewicz, R.~W. Blair, C.~A. Moodie, A.~I. Isik, A.~Erramuzpe, J.~D. Kent, M.~Goncalves, E.~DuPre, M.~Snyder \emph{et~al.}, ``fmriprep: a robust preprocessing pipeline for functional mri,'' \emph{Nature methods}, vol.~16, no.~1, pp. 111--116, 2019.

\bibitem{marcus2011informatics}
D.~S. Marcus, J.~Harwell, T.~Olsen, M.~Hodge, M.~F. Glasser, F.~Prior, M.~Jenkinson, T.~Laumann, S.~W. Curtiss, and D.~C. Van~Essen, ``Informatics and data mining tools and strategies for the human connectome project,'' \emph{Frontiers in neuroinformatics}, vol.~5, p.~4, 2011.

\bibitem{li2022atlas}
Y.~Li, A.~Liu, X.~Fu, M.~J. Mckeown, Z.~J. Wang, and X.~Chen, ``Atlas-guided parcellation: Individualized functionally-homogenous parcellation in cerebral cortex,'' \emph{Computers in Biology and Medicine}, vol. 150, p. 106078, 2022.

\end{thebibliography}

\end{document}